\documentclass[10pt,twocolumn,letterpaper]{article}

\usepackage[pagenumbers]{cvpr} 

\definecolor{cvprblue}{rgb}{0.21,0.49,0.74}
\usepackage[pagebackref,breaklinks,colorlinks,allcolors=cvprblue]{hyperref}
\usepackage{multirow}

\def\confName{CVPR}
\def\confYear{2027}

\title{Beyond Episodic Recall: Experience-to-Knowledge Evolution for Lifelong Embodied Navigation}

\author{Weitao An\\
Xidian University\\
Xi'an, China
\and
Chenghao Xu\\
Hohai University\\
Nanjing, China
\and
Xu Yang\\
Hohai University\\
Nanjing, China
\and
Cheng Deng\\
Xidian University\\
Xi'an, China
}

\begin{document}
\maketitle
\begin{abstract}

Lifelong embodied navigation requires agents to accumulate experience through continuous interaction and reuse it for subsequent goal searches. Existing memory-enhanced methods typically retain historical observations or construct structured scene representations. However, storing observations or high-dimensional semantic embeddings incurs growing storage costs. Meanwhile, structured representations support retrieval and semantic reasoning but do not explicitly evolve accumulated experience into environment-specific knowledge. To address these limitations, we propose Experience-Evolving Memory (EvoNav), a novel two-timescale memory framework.  It organizes semantic observations, spatial relations, and task-feedback cues into sparse experience atoms. During online operation, experience atoms continually consolidate repeated observations and form dynamic links based on transitions. At the slower timescale, EvoNav distills accumulated category co-occurrences and feedback signals into environment-specific structural knowledge without relying on predefined object--room relations. For each new goal, EvoNav combines direct goal evidence from experience atoms with learned structural knowledge to guide goal-directed search on the current traversability map. Experiments on single-goal, sequential multi-floor, and multimodal navigation demonstrate effective experience reuse and improved path efficiency with compact persistent memory, together with successful transfer to a real robot.

\end{abstract}    

\section{Introduction}
\label{sec:evonav_intro}
\begin{figure}[ht]
    \centering
    \includegraphics[width=\linewidth]{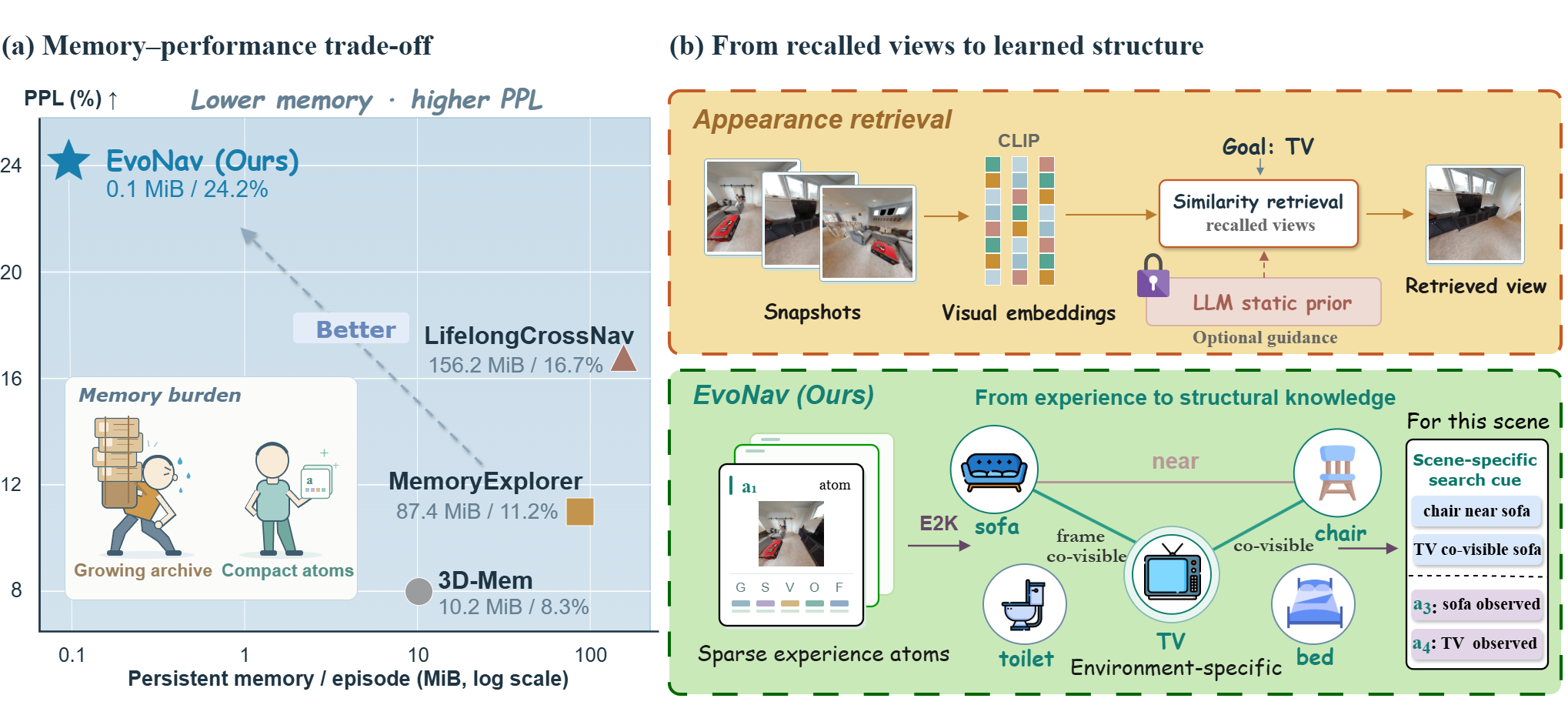}
    \caption{\textbf{Motivation of EvoNav.}
    (a) The trade-off between core persistent semantic-memory footprint and partial path length, which weights partial task progress by the path efficiency of successful subtasks, on MFMON.
    (b) Appearance-recall memories retain snapshots or high-dimensional visual
    features and return previously observed views, whereas EvoNav consolidates
    interaction histories into sparse experience atoms and evolves
    environment-specific structural knowledge.}
    \label{fig:evonav_motivation}
\end{figure}
Zero-shot ObjectNav systems combine semantic mapping with goal-directed exploration to search unfamiliar scenes~\cite{zhou2023esc,yu2023l3mvn,yokoyama2024vlfm}. Practical agents, however, often execute successive goals in the same multi-floor environment, where past observations and traversal can inform later search. This calls for accumulating reusable knowledge across tasks~\cite{wani2020multion,chang2024goat}.

Existing memory-enhanced navigation methods represent past experience in several forms. Image-based methods directly retain visual snapshots or multi-view observations for later retrieval~\cite{yang2025threedmem,wang2026lmee}. Feature-based methods project language-aligned semantic embeddings into 3D space, producing spatial feature maps that support open-vocabulary queries~\cite{huang2023vlmaps,peng2023openscene,jatavallabhula2023conceptfusion}. Other methods adopt structured representations, organizing perceived entities, spatial topology, and interaction history into object-relation, hierarchical, or persistent memory graphs for semantic reasoning and decision-making~\cite{gu2024conceptgraphs,rana2023sayplan,werby2024hovsg,niu2026ssmgnav,chen2026stegnav}. In multi-floor environments, 3D traversability modeling and floor connectivity further make cross-floor navigation decisions executable~\cite{zheng2026travexplorer,li2026lifelongcrossnav}. Together, these directions establish important foundations for experience reuse, structured reasoning, and cross-floor sequential navigation.

Despite this progress, a gap remains, as illustrated in Figure~\ref{fig:evonav_motivation}, between lightweight experience accumulation and learning environment-specific structural knowledge. Accumulating visual observations and high-dimensional spatial features increases storage and maintenance costs as interaction grows~\cite{yang2025threedmem}, but their immediate value is often limited to relocating a previously observed target. Scene graphs are commonly designed to organize perceptual entities or provide task-time relational context, focusing on semantic reasoning and decision-making rather than distilling repeated traversal and task outcomes into compact, feedback-calibrated memory~\cite{gu2024conceptgraphs,yin2024sgnav,yin2025unigoal,huang2026msgnav}. Likewise, external commonsense priors can guide exploration~\cite{zhou2023esc,yu2023l3mvn}, but may not reflect the structure of the current environment~\cite{zhang2023layout}. In contrast, repeated object co-occurrences, behavior-supported transitions, and task feedback provide direct signals for learning how a particular environment is organized. We therefore ask: \emph{how can an agent retain reusable navigation experience compactly, and evolve it into environment-specific knowledge that guides subsequent searches?} Inspired by two-timescale learning~\cite{mahadevan2026convergence}, we formulate a memory-learning paradigm coupling fast online consolidation with slow environment-specific knowledge evolution.

Accordingly, we propose \textbf{EvoNav}, a two-timescale experience-to-knowledge framework for lifelong embodied navigation. Drawing on the atomic notes and dynamic links of Zettelkasten~\cite{kadavy2021digitalzettelkasten,xu2025amem}, EvoNav represents local interactions as sparse experience atoms containing semantic, spatial, relational, and feedback cues without imposing a fixed object=room hierarchy. At the fast timescale, \emph{online experience consolidation} merges repeated observations into existing atoms and maintains perceptual and behavior-supported transition links. At the slow timescale, \emph{experience-to-knowledge evolution} transforms these atom-level relations into environment-specific category-level co-occurrence and transition patterns, while task outcomes calibrate the goal-conditioned reliability of each atom. For each new goal, \emph{Atom=Graph Evidence Propagation} (AGEP) combines direct atom evidence with learned structural support. The fused evidence is then propagated over perceptual and transition links to rank relevant atoms. Atom=object binding and traversability checks then ground the selected atoms into executable navigation targets.

Our main contributions are as follows:
\begin{itemize}
    \item \textbf{Sparse experience atom memory.}
    We introduce sparse experience atoms that compactly encode navigation-relevant semantic, spatial, and feedback cues, without relying on a fixed object-room hierarchy or persistently storing visual snapshots and per-voxel high-dimensional semantic features.

    \item \textbf{Memory evolution across timescales.}
    We coordinate online experience consolidation with slower experience-to-knowledge evolution, integrating repeated observations into experience and learning environment-specific structural knowledge from co-occurrences, behavioral transitions, and task feedback for subsequent searches.

    \item \textbf{Experience-graph reasoning and grounded navigation.}
    We combine direct evidence with learned structural support, using atom-ObjectMap binding and feasibility checks on the current traversability map to turn historical experience into actionable navigation candidates across tasks and floors.

    \item \textbf{Comprehensive evaluation and real-world deployment.}
    Experiments on single-goal, sequential multi-floor, and multimodal navigation demonstrate strong path efficiency with a compact persistent-memory footprint. Deployment on a real mobile robot further validates the feasibility of the system.
    
\end{itemize}

\section{Related Work}
\label{sec:related_work}

\subsection{Goal Search and Cross-Floor Execution}

Map-based ObjectNav combines goal-directed exploration with global-local planning, semantic maps, and learned objectives~\cite{chaplot2020semantic,ramakrishnan2022poni}. End-to-end policies and topological memories instead learn search behavior or reason over previously visited observations~\cite{chaplot2020topological}. CoW, VLFM, ZSON, and GAMap use open-vocabulary evidence, multimodal goal embeddings, or geometric affordances~\cite{gadre2023cows,yokoyama2024vlfm,majumdar2022zson,yuan2024gamap}, whereas ESC and L3MVN score unexplored regions with pretrained object-scene priors~\cite{zhou2023esc,yu2023l3mvn}. For multi-floor navigation, TravExplorer and LifelongCrossNav further model 3-D traversability, stairs, and cross-floor paths~\cite{zheng2026travexplorer,li2026lifelongcrossnav}.

\subsection{Open-Vocabulary Maps and Scene Graphs}

Open-vocabulary maps attach language-aligned visual features to spatial locations, enabling post-hoc scene queries through dense, point-level, or implicit field representations~\cite{huang2023vlmaps,peng2023openscene,jatavallabhula2023conceptfusion}. Scene graphs instead organize environments around entities and relations. Early 3-D scene graphs unify objects, rooms, cameras, and spatial relations, while dynamic scene graphs add actionable topology and moving agents~\cite{armeni2019scenegraph,rosinol2020dsg}. Hydra constructs hierarchical 3-D scene graphs online, and ConceptGraphs, OVSG, and Open3DSG support open-vocabulary objects, contextual entities, and relations~\cite{gu2024conceptgraphs,chang2023ovsg,koch2024open3dsg}. Such structures can also support planning through task-relevant subgraph retrieval, online scene-graph prompting, floor-room-object hierarchies, or multimodal graph selection~\cite{rana2023sayplan,yin2024sgnav,werby2024hovsg,huang2026msgnav}. These representations offer complementary trade-offs between semantic coverage, organization, and scene information.

\subsection{Sequential Navigation and Experience Memory}
MultiON, GOAT, GOAT-Bench, and LMEE-Bench evaluate accumulated memory across multi-object, multimodal sequential-navigation, exploration, and question-answering settings~\cite{wani2020multion,chang2024goat,khanna2024goatbench,wang2026lmee}. Navigation memories encode observation-action histories and states or retain goal-relevant working memory over topological maps~\cite{fang2019smt,pashevich2021episodic,du2023hinl,li2024memonav}. Other methods learn reusable abstractions, retrieve episodic experience, compress snapshots or trajectories, or update memory from feedback~\cite{campari2022abstract,wang2022vienna,ziliotto2025tango,dai2026history,yang2025threedmem,hu2026astranav,wang2026trajrag,ge2026evomemnav}. ReMEmbR builds retrieval-augmented spatio-temporal memory, while LifelongCrossNav maintains 3-D semantic voxels for sequential cross-floor goals~\cite{anwar2024remembr,li2026lifelongcrossnav}.

\begin{figure*}[ht]
    \centering
    \includegraphics[width=\textwidth]{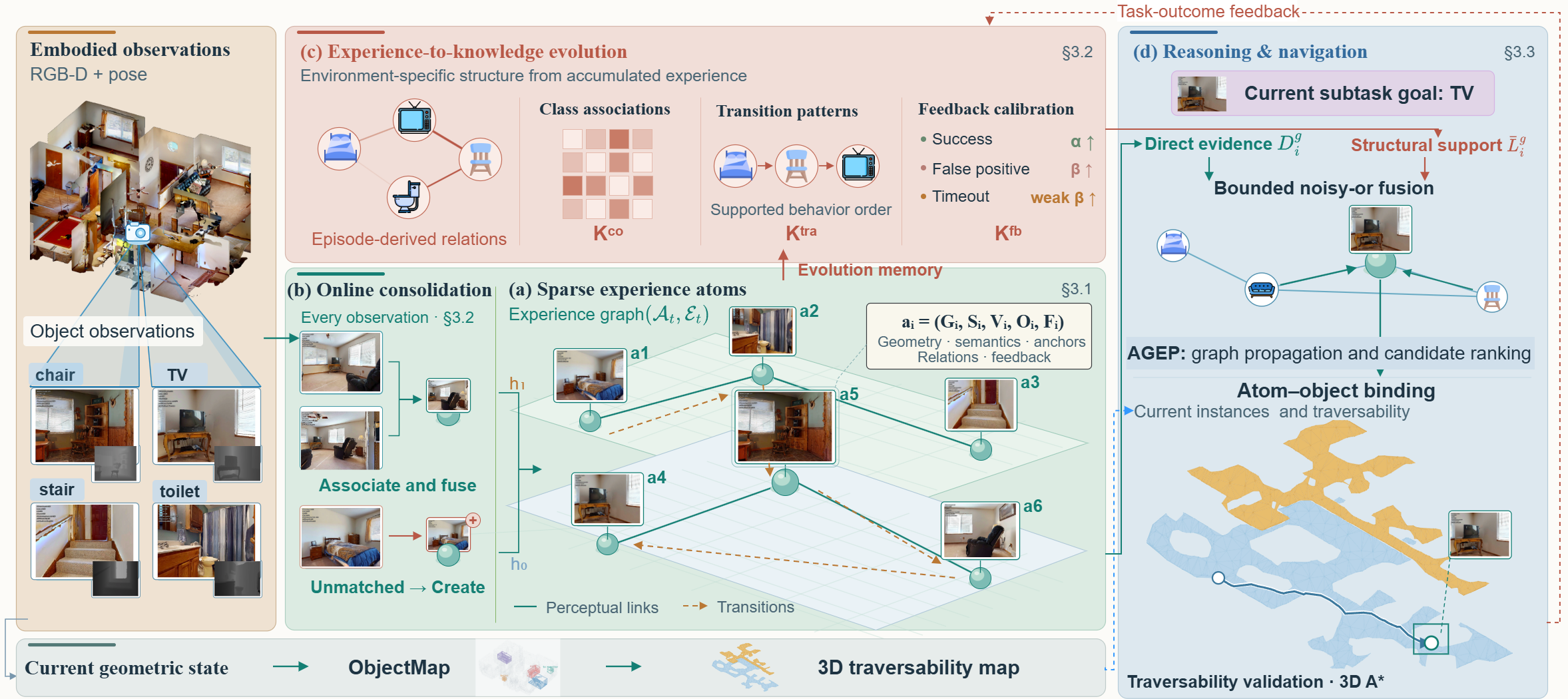}
    \caption{\textbf{Overview of EvoNav.} (a) Sparse experience atoms retain compact
    geometry, semantics, observation anchors, relations, and feedback.
    (b) Online consolidation associates repeated observations with existing atoms
    and creates new atoms for unmatched experience. (c) Experience-to-Knowledge
    evolution extracts environment-specific class associations, transition
    patterns, and feedback-calibrated reliability. (d) AGEP fuses direct evidence
    with structural support, ranks candidate atoms, grounds them in current
    ObjectMap instances, and constrains planning using the 3-D traversability map.}
    \label{fig:evonav_overview}
\end{figure*}

\section{Method}

We propose EvoNav, which organizes navigation observations and interaction feedback into sparse atomic experience. As shown in Figure~\ref{fig:evonav_overview}, EvoNav has two core modules. The first, illustrated in Fig.~\ref{fig:evonav_overview}(a)-(c), performs two-timescale \emph{Experience-to-Knowledge evolution} by retaining stable observations through online consolidation and extracting environment-specific structural knowledge at a slower timescale. The second, shown in Fig.~\ref{fig:evonav_overview}(d), uses \emph{Atom-Graph Evidence Propagation} (AGEP) to combine direct experience evidence with learned structural knowledge and retrieve goal-relevant atoms. These atoms are then grounded into executable navigation targets through instance binding and traversability checks. At time $t$, the atomic experience memory is:
\begin{equation}
\mathcal M_t=(\mathcal A_t,\mathcal E_t,\mathcal K_t),
\end{equation}
where $\mathcal A_t$ is the set of atoms, $\mathcal E_t$ contains experience links, and $\mathcal K_t$ represents reusable scene knowledge derived within the current episode. Atoms and experience links are updated online, while category-level summaries and feedback calibration are refreshed at subtask boundaries. The system separately maintains a compact ObjectMap and a goal-independent traversability map. The former provides instance-level geometry for grounding, whereas the latter uses obstacle information to validate candidate positions and support path planning.

\subsection*{3.0 Problem Definition}

Given a goal sequence $(g_1,\ldots,g_K)$ in the same unknown environment, the agent receives $g_k$ only when the current subtask starts and cannot access future goals during earlier subtasks. At time $t$, it selects an action using RGB-D observations, its current pose, and accumulated memory. It searches for the current goal, declares that the goal has been found, and moves to the next subtask according to the task protocol. Maps and memory are retained across subtasks and reset at the start of a new episode. 

\subsection{Sparse Experience Representation}

Our system builds a compact geometric ObjectMap and an atomic experience memory in parallel from the same RGB-D stream. Object masks are projected into 3D, clustered, and associated across frames to update ObjectMap instances with compact spatial extent, multi-class confidence, and observation support. A separate goal-independent traversability map supports candidate validation and path planning. Inspired by Zettelkasten~\cite{kadavy2021digitalzettelkasten,ahrens2017smartnotes,xu2025amem}, a note-taking method that organizes experience as interconnected atomic notes, we consolidate object observations and their interaction context into sparse experience atoms. Each atom retains compact spatial summaries, semantic evidence, past observation positions, and task feedback. The ObjectMap provides compact instance-level geometry for grounding, whereas atom memory accumulates reusable historical experience for knowledge evolution and later search. Specifically, an atom is represented as:
\begin{equation}
a_i=(\mathcal G_i,\mathcal S_i,\mathcal V_i,\mathcal O_i,\mathcal F_i).
\label{eq:evonav_atom}
\end{equation}
The spatial support $\mathcal G_i=(\mathbf c_i,\mathbf B_i,h_i)$ contains a 3D center estimated from object point clouds, a bounding box, and a height-level index. Semantic evidence takes a sparse multi-label form:
\begin{equation}
\mathcal S_i=\{(y,s_i^y,n_i^y)\}_{y\in\mathcal Y_i},
\qquad
q_i^y=\frac{s_i^y}{\sum_{u\in\mathcal Y_i}s_i^u+\epsilon},
\end{equation}
where $\mathcal Y_i$ is the set of classes recorded by the atom, $s_i^y$ is the fused detection confidence, $n_i^y$ is the observation count, and $\epsilon$ is a small constant for numerical stability. The normalized evidence $q_i^y$ measures support for a class relative to other interpretations, retaining class ambiguity across repeated observations.

$\mathcal V_i$ stores a bounded set of past observation positions and observation metadata, while $\mathcal O_i$ records local object proximity and same-frame co-visibility. The feedback state $\mathcal F_i$ records successes, false-positive stops, and timeouts. To characterize the historical reliability of atom $i$ for goal $g$, we maintain a goal-conditioned Beta belief and define the direct memory evidence as:
\begin{equation}
\theta_i^g\sim\operatorname{Beta}(\alpha_i^g,\beta_i^g),
\qquad
D_i^g=q_i^g\mathbb{E}[\theta_i^g]
=q_i^g\frac{\alpha_i^g}{\alpha_i^g+\beta_i^g}.
\label{eq:evonav_direct}
\end{equation}
Here, $\alpha_i^g$ and $\beta_i^g$ accumulate positive and negative feedback evidence, respectively, and are updated from subtask outcomes using soft evidence counts. Atoms store compact geometry, class evidence, and sparse relations without keeping high-dimensional vision-language embeddings at every voxel. The ObjectMap retains the instance geometry.

\subsection{Two-Timescale Experience-to-Knowledge Evolution}

This module updates experience memory at two timescales. The online stage continuously absorbs and compresses local observations, while the slower stage evolves accumulated experience, links, and feedback into reusable environment-specific structural knowledge at subtask boundaries.

\paragraph{Online Experience Consolidation.}

RGB-D perception produces an object observation $o_t=(\mathbf c_t,\mathbf B_t,y_t,s_t,\mathbf v_t)$, which contains the point cloud center, bounding box, class, detection confidence, and observation position. A two-level association rule merges new observations into existing atoms. Strong association relies on bounding-box overlap or close spatial
agreement. A broader weak association is allowed only for
semantically compatible observations. A new atom is created when no valid association exists.

Repeated observations consolidate geometry and semantics through incremental updates:
\begin{equation}
\mathbf c_i^+=\frac{m_i\mathbf c_i+\mathbf c_t}{m_i+1},
\qquad
s_i^{y_t,+}=\frac{n_i^{y_t}s_i^{y_t}+s_t}{n_i^{y_t}+1},
\end{equation}
where $m_i$ is the total observation count. A newly observed class is initialized with its current confidence. The bounding box is updated to the axis-aligned box enclosing past and current observations, and past observation positions are stored.

The online stage also maintains:
\begin{equation}
\mathcal E_t=\mathcal E_t^{\mathrm{per}}\cup\mathcal E_t^{\mathrm{tra}}.
\end{equation}
At the graph level, the atom-local relation records in $\mathcal O_i$
form $\mathcal E_t^{\mathrm{per}}$, with repeated observations
increasing their support counts. Transition links
$\mathcal E_t^{\mathrm{tra}}$ are formed from successive atom
activations that satisfy movement and time-gap constraints, and record the source, destination, and transition distance. These links retain perceptual relations and behavioral order, forming an experience structure that changes with interaction.

\paragraph{Environment-Specific Knowledge Evolution.}

At the end of each subtask, slow memory combines consolidated observations, experience links, and execution feedback into:
\begin{equation}
\mathcal K_t=(\mathcal K_t^{\mathrm{co}},\mathcal K_t^{\mathrm{tra}},\mathcal K_t^{\mathrm{fb}}),
\end{equation}
which respectively encode category associations, transition links
$\mathcal E_t^{\mathrm{tra}}$ used in graph propagation, and
feedback-calibrated atom reliabilities $R_i^g$. All are derived from
experience within the current episode without predefined
class-to-room rules.

For classes $u,v$, let $N$ be the current number of atoms, $N_u$ the number
containing class $u$, and $N_{uv}$ the number of distinct atoms supporting
relation $(u,v)$. We define $\widehat P_t(u)=N_u/N$ and
$\widehat P_t(u,v)=N_{uv}/N$, and measure their association relative to
independent occurrence using a positive pointwise mutual information form:
\begin{equation}
I_t(u,v)=\left[\log\frac{\widehat P_t(u,v)+\epsilon}
{\max(\widehat P_t(u)\widehat P_t(v),\epsilon)}\right]_+,
\end{equation}
where $[x]_+=\max(x,0)$.
Information gain alone does not indicate whether a relation has enough observation support. To turn it into scene knowledge for later retrieval, we combine repeated support, class specificity, and relation source quality into a bounded predictive association score:
\begin{equation}
\kappa_t(u,v)=(1-e^{-I_t(u,v)})(1-e^{-n_{uv}})\xi_{uv}\rho_{uv},
\label{eq:evonav_layout_relation}
\end{equation}
where $n_{uv}$ is the relation support count. Class specificity $\xi_{uv}$ measures spatial distinctiveness based on class prevalence in the current memory, while $\rho_{uv}$ measures evidence quality based on relation sources. $\kappa_t(u,v)$ measures how strongly observing $u$ supports prioritizing its associated region when searching for $v$. This form reduces the influence of accidental co-occurrence and common classes. The predictive association scores serve as edge weights in the class knowledge:
\begin{equation}
\mathcal K_t^{\mathrm{co}}=\{(u,v,\kappa_t(u,v))\}.
\end{equation}
For a query goal $g$, if a context class $y$ recorded by atom $a_i$ has an experience-supported association with $g$, then $\kappa_t(y,g)$ provides an indirect search cue. We combine these class cues with local object relations, transition patterns, and feedback into goal-specific structural support:
\begin{equation}
L_i^g=\Psi(a_i,g,\mathcal K_t),
\label{eq:evonav_layout_support}
\end{equation}
where $\Psi$ is a fixed bounded evidence aggregation function over class associations, local relations, transition context, task feedback, and distance to historical anchors. 

Task feedback increases $\alpha_i^g$ after success and $\beta_i^g$ after a false positive, while timeouts and other failures receive weaker negative feedback. Slow evolution both calibrates experience reliability and extracts scene structure, allowing later queries to use accumulated experience.

\begin{figure}[ht]
    \centering
    \includegraphics[width=\linewidth]{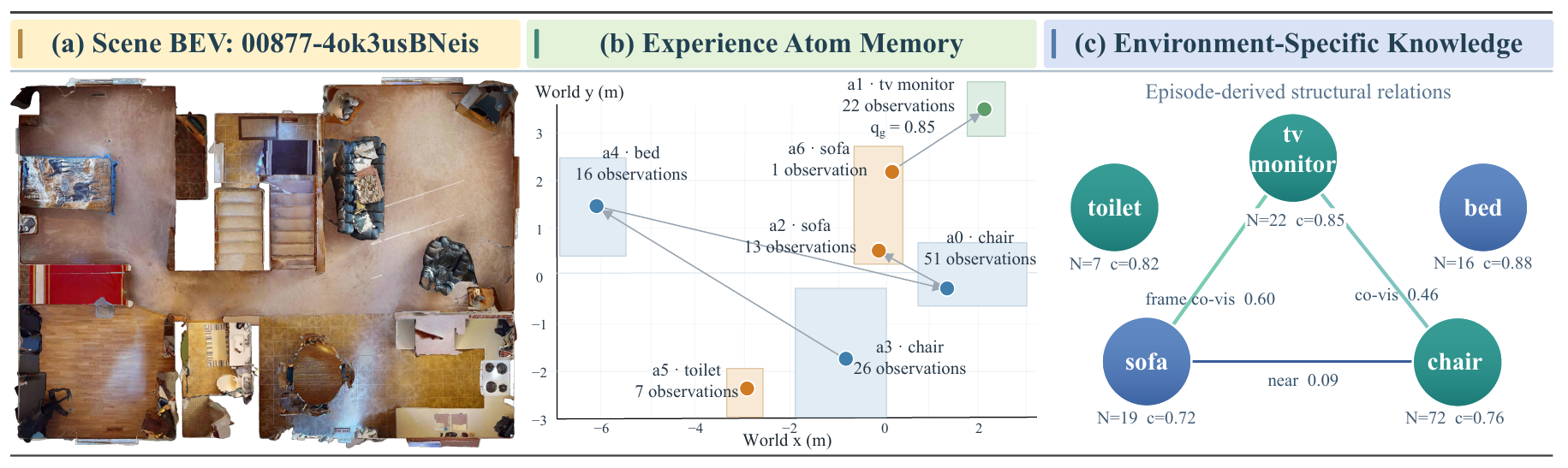}
    \caption{\textbf{Experience-to-Knowledge evolution in an explored scene.}
    Left: a bird's-eye view of the observed HM3D scene. Middle: consolidated
    atoms in the shared world coordinate system. Right: category-level
    structural knowledge evolved from atom relations, including proximity,
    atom-level co-visibility, and frame-level category co-visibility.}

    \label{fig:atom_knowledge_example}
\end{figure}

As illustrated in Fig.~\ref{fig:atom_knowledge_example}, repeated scene observations are first consolidated into sparse atoms in a shared world coordinate system and subsequently evolved, through recurring atom relations, into category-level environment-specific structural knowledge.

\subsection{AGEP-Guided Geometric Grounding}

\paragraph{Atom-Graph Evidence Propagation.}

When a new goal $g$ arrives, retrieval considers both direct goal observations and indirect structural cues. We refer to this goal-conditioned evidence initialization, graph propagation, and candidate ranking process as Atom-Graph Evidence Propagation (AGEP). Specifically, we calibrate $L_i^g$ using the goal-conditioned atom reliability and bound its magnitude to obtain effective structural evidence $\bar L_i^g$. For atoms with clear goal observations, we further limit the structural increment to reduce context interference in choosing among instances of the same class. We then fuse it with direct evidence $D_i^g$ using noisy-or:
\begin{equation}
E_i^{(0),g}=1-(1-D_i^g)(1-\bar L_i^g).
\label{eq:evonav_initial}
\end{equation}
We further apply bounded evidence propagation over the experience graph. Let $\mathcal N(i)$ be the neighbors linked to atom $i$, $w_{ji}\in[0,1]$ the coefficient for support collected from related experience $j$ into atom $i$, and $k$ the propagation round. Then:
\begin{align}
C_i^{(k),g}&=1-\prod_{j\in\mathcal N(i)}(1-w_{ji}E_j^{(k),g}),\\
\bar C_i^{(k),g}&=\min\left(C_i^{(k),g},c_{\mathrm{ctx}}\max_j E_j^{(0),g}\right),\\
E_i^{(k+1),g}&=1-(1-E_i^{(0),g})(1-\bar C_i^{(k),g}).
\label{eq:evonav_agep}
\end{align}
Here, $C_i^{(k),g}$ is aggregated neighbor support, $\bar C_i^{(k),g}$ is bounded support, and $c_{\mathrm{ctx}}$ controls the context bound. After $H$ rounds, we obtain $E_i^g=E_i^{(H),g}$. Atoms whose goal-class support meets the minimum threshold are ranked primarily by direct evidence, while indirect candidates use propagated evidence. Both types come from atomic experience memory. 
\paragraph{Atom-ObjectMap Grounding and Navigation.}

AGEP outputs a goal-relevant atom ranking but does not directly declare success or produce control actions. Retrieved atoms are subsequently bound to instance clusters of the current goal class in the ObjectMap. For $o_j=(\mathbf c_j,\mathbf B_j,p_j^g)$, the terms denote the instance center, bounding box, and goal-class confidence. Geometric and semantic agreement is:
\begin{equation}
\ell_{ij}^g=p_j^gE_i^g
\exp\left(-\frac{d_{xy}^2(i,j)}{2\sigma_{xy}^2}-\frac{d_z^2(i,j)}{2\sigma_z^2}\right)
\frac{1+\operatorname{IoU}(\mathbf B_i,\mathbf B_j)}{2},
\label{eq:evonav_binding_likelihood}
\end{equation}
where $d_{xy}(i,j)$ and $d_z(i,j)$ are the horizontal distance and absolute height difference between the atom and instance centers. $\sigma_{xy}$ and $\sigma_z$ are spatial matching scales, and $\operatorname{IoU}$ is 3D bounding box overlap. Since $\ell_{ij}^g\in[0,1]$, we map it to a bounded score without introducing additional calibration parameters:
\begin{equation}
B_{ij}^g=\operatorname{sigmoid}(\ell_{ij}^g),
\qquad
\operatorname{sigmoid}(z)=\frac{1}{1+e^{-z}}.
\label{eq:evonav_binding_score}
\end{equation}
This monotonic mapping preserves the likelihood ordering of candidates. Let $\mathcal F_t^g$ denote the set of atom-instance candidates that pass the goal-class, spatial, minimum-binding-score, instance-confidence, and current-map reachability checks. Let $\operatorname{rank}_g(i)$ be the AGEP retrieval rank of atom $i$, and let $\mathbb I_{\mathrm{sf}}(j)$ indicate whether instance $j$ is on the same floor as the agent. The grounded target is selected by lexicographic maximization:
\begin{equation}
(i^\star,j^\star)=
\operatorname*{arg\,max}^{\mathrm{lex}}_{(i,j)\in\mathcal F_t^g}
\left(\mathbb I_{\mathrm{sf}}(j),B_{ij}^g,-\operatorname{rank}_g(i)\right).
\label{eq:evonav_grounded_selection}
\end{equation}
The planner samples landing positions around instance $j^\star$, validates them on the obstacle-inflated traversability map, and plans with 3D A*. If no executable instance binding exists, it tries past observation anchors in AGEP order; reaching an anchor triggers revisiting rather than success. When all memory candidates are infeasible, the agent returns to frontier exploration.

\section{Experiments}
\label{sec:experiments}

\subsection{Experimental Setup}
\label{sec:exp_setup}

\paragraph{Benchmarks.}
We evaluate on three complementary benchmarks. HM3D ObjectNav isolates
single-goal navigation on 2,000 validation episodes from 20 scenes;
HM3D-MFMON contains 927 ordered three-goal episodes from 36 multi-floor
scenes, including 288 Cross-Floor-Required episodes, and is our primary test of
long-horizon and cross-floor experience reuse \cite{li2026lifelongcrossnav};
GOAT-Bench evaluates 5-10 category, language-description, and
instance-image goals on the Val-Unseen split
\cite{khanna2024goatbench}.

\paragraph{Setup.}
The agent receives observations from a $640\times480$ RGB-D camera with a depth
range of $[0,5.0]$ m and has a radius of $0.18$ m. Its discrete action space
contains \textsc{MoveForward} ($0.25$ m), $30^{\circ}$ left/right turns,
$45^{\circ}$ camera look-up/look-down actions, and \textsc{Stop}. Each subtask
has a budget of 500 actions, and a \textsc{Stop} is considered successful when
the agent is within $1.0$ m of the target. The main text reports the prefix protocol, where an episode terminates at its first failed subtask.

\paragraph{Evaluation metrics.}
For single-goal navigation, we report success rate (SR) and success weighted by
path length (SPL). SR measures the fraction of successful episodes, while SPL
additionally evaluates path efficiency using the ratio between geodesic shortest
and executed path lengths. For MFMON, sequence success rate (SeqSR) requires all
ordered goals in an episode to succeed, and sequence SPL (SeqSPL) evaluates the
path efficiency of the complete sequence under this strict success condition.
Partial success rate (PR) is the mean fraction of completed subtasks per episode,
whereas partial path length (PPL) weights this partial progress by the path
efficiency of successful subtasks.

\subsection{Main Results}
\label{sec:main_results}

\paragraph{Single-goal ObjectNav.}
Table~\ref{tab:hm3d_objectnav} reports the single-goal ObjectNav results on HM3D.
After introducing atomic memory, knowledge evolution, and memory grounding,
EvoNav remains competitive in success rate while retaining strong path efficiency,
showing that the memory extension preserves the underlying capabilities for target
perception, cross-floor search, and geometric planning. The long-horizon experiments
further evaluate the benefit of reusing accumulated experience.

\begin{table}[t]
\centering
\caption{HM3D ObjectNav validation results (\%). TF denotes training-free with
respect to task-specific navigation data, and LLM denotes the use of an LLM for semantic reasoning.}
\label{tab:hm3d_objectnav}
\small
\begin{tabular}{lcccc}
\toprule
Method & TF & LLM & SR $\uparrow$ & SPL $\uparrow$ \\
\midrule
L3MVN (2023)~\cite{yu2023l3mvn} & \checkmark & \checkmark & 50.4 & 23.1 \\
VLFM (2024)~\cite{yokoyama2024vlfm} & \checkmark & $\times$ & 52.5 & 30.4 \\
ApexNav (2025)~\cite{zhang2025apexnav} & \checkmark & \checkmark & 59.6 & 33.0 \\
MFNP (2025)~\cite{zhang2025mfnp} & \checkmark & \checkmark & 58.3 & 26.7 \\
ASCENT (2026)~\cite{gong2026ascent} & \checkmark & \checkmark & 65.4 & 33.5 \\
SR-Nav (2026)~\cite{fang2026srnav} & \checkmark & \checkmark & 58.3 & 33.0 \\
\midrule
EvoNav (ours) & \checkmark & $\times$ & \textbf{65.9} & \textbf{39.7} \\
\bottomrule
\end{tabular}
\end{table}

\paragraph{Ordered multi-floor navigation.}
As shown in Table~\ref{tab:mfmon_all}, EvoNav achieves $30.31\%$ SeqSR,
$15.54\%$ SeqSPL, $48.26\%$ PR, and $24.17\%$ PPL over all 927 MFMON
episodes. Compared with the strongest published baseline, LifelongCrossNav,
EvoNav improves SeqSR, SeqSPL, and PPL by $1.18$, $5.90$, and $7.47$
percentage points, respectively. These results suggest that EvoNav's
primary advantage lies in path efficiency and later-goal execution.
The contribution of experience reuse to these gains is further examined
through the ablation studies.

\begin{table}[t]
\centering
\caption{Results on all 927 HM3D-MFMON episodes under the prefix protocol (\%).
SeqSR/SeqSPL are strict episode-level metrics, whereas PR/PPL measure goal-level
partial completion within each episode.}
\label{tab:mfmon_all}
\small
\begin{tabular}{lcccc}
\toprule
Method & SeqSR $\uparrow$ & SeqSPL $\uparrow$ & PR $\uparrow$ & PPL $\uparrow$ \\
\midrule
OneMap & 16.83 & 6.95 & 33.69 & 14.03 \\
3D-Mem & 8.31 & 3.62 & 22.14 & 8.27 \\
MemoryExplorer  & 12.19 & 5.73 & 27.26 & 11.24 \\
LifelongCrossNav & 29.13 & 9.64 & \textbf{48.40} & 16.70 \\
\midrule
EvoNav (ours) & \textbf{30.31} & \textbf{15.54} & 48.26 & \textbf{24.17} \\
\bottomrule
\end{tabular}
\end{table}

\begin{table*}[t]
\centering
\caption{Position-wise Global SR/SPL on MFMON under the prefix protocol (\%).
All positions use 927 episodes as the denominator, with unexecuted later goals
assigned zero.}
\label{tab:mfmon_position}

\begin{tabular}{lcccccc}
\toprule
& \multicolumn{2}{c}{Subtask 0} & \multicolumn{2}{c}{Subtask 1} &
\multicolumn{2}{c}{Subtask 2} \\
\cmidrule(lr){2-3}\cmidrule(lr){4-5}\cmidrule(lr){6-7}
Method & SR $\uparrow$ & SPL $\uparrow$ & SR $\uparrow$ & SPL $\uparrow$ & SR $\uparrow$& SPL $\uparrow$ \\
\midrule
OneMap & 54.58 & 23.46 & 29.67 & 13.54 & 16.83 & 9.14 \\
3D-Mem  & 38.08 & 17.36 & 20.06 & 9.21 & 8.31 & 4.63 \\
MemoryExplorer  & 43.04 & 21.27 & 26.32 & 14.18 & 12.19 & 8.09 \\
LifelongCrossNav w/o H-POI & \textbf{70.33} & 25.26 & 45.52 & 18.06 & 29.77 & 11.64 \\
LifelongCrossNav & \textbf{70.33} & 25.26 & 45.74 & 22.71 & 29.13 & 15.82 \\
\midrule
EvoNav (ours) & $68.18$ & $\mathbf{35.03}$ &
$\mathbf{46.28}$ & $\mathbf{26.19}$ &
$\mathbf{30.31}$ & $\mathbf{18.86}$ \\
\bottomrule
\end{tabular}
\end{table*}

\paragraph{Position-wise performance and experience reuse.}
Table~\ref{tab:mfmon_position} reports position-wise Global SR/SPL under the
prefix protocol. Every position uses all 927 episodes as its denominator, and
unexecuted goals after an earlier failure receive zero. EvoNav obtains the best
Global SPL at all three positions: $35.03\%$, $26.19\%$, and $18.86\%$. EvoNav's Subtask~0 Global SR is lower than LifelongCrossNav by
$2.15$ points, but it closes this gap and surpasses LifelongCrossNav at
Subtasks~1 and~2, reaching the highest SR of $46.28\%$ and $30.31\%$,
respectively. The results show that its main advantage
lies in path efficiency and later-goal completion.

Table~\ref{tab:mfmon_floor_split} separates CFR from non-CFR episodes. On CFR, 
EvoNav improves SeqSR and SeqSPL over LifelongCrossNav by $0.69$ and $0.73$ 
percentage points, respectively, and raises PPL by $3.41$ points. On non-CFR, EvoNav obtains the best SeqSR, SeqSPL, 
and PPL, exceeding LifelongCrossNav by $1.40$ SeqSR points, $8.23$ SeqSPL points, 
and $9.30$ PPL points. The path-efficiency 
advantage is larger on non-CFR episodes, while the low absolute SeqSR on CFR 
episodes highlights the remaining challenge of cross-floor navigation.

\begin{table}[t]
\centering
\caption{Cross-floor decomposition on HM3D-MFMON under the prefix protocol
(\%). }
\label{tab:mfmon_floor_split}
\small
\begin{tabular}{lcccc}
\toprule
Method & SeqSR $\uparrow$ & SeqSPL $\uparrow$ & PR $\uparrow$ & PPL $\uparrow$ \\
\midrule
\multicolumn{5}{c}{\textbf{(a) Cross-Floor-Required: 288 episodes}} \\
\midrule
OneMap & 0.00 & 0.00 & 18.52 & 7.97 \\
LifelongCrossNav & 7.99 & 2.35 & \textbf{29.17} & 9.75 \\
\midrule
EvoNav (ours) & \textbf{8.68} & $\mathbf{3.08}$ & 28.01 & $\mathbf{13.16}$ \\
\midrule
\multicolumn{5}{c}{\textbf{(b) Non-Cross-Floor: 639 episodes}} \\
\midrule
OneMap & 24.42 & 10.08 & 40.53 & 16.76 \\
LifelongCrossNav & 38.66 & 12.93 & 57.07 & 19.83 \\
\midrule
EvoNav (ours) & $\mathbf{40.06}$ & $\mathbf{21.16}$ & \textbf{57.38} & $\mathbf{29.13}$ \\
\bottomrule
\end{tabular}
\end{table}

\begin{figure}[t]
    \centering
    \includegraphics[width=\linewidth]{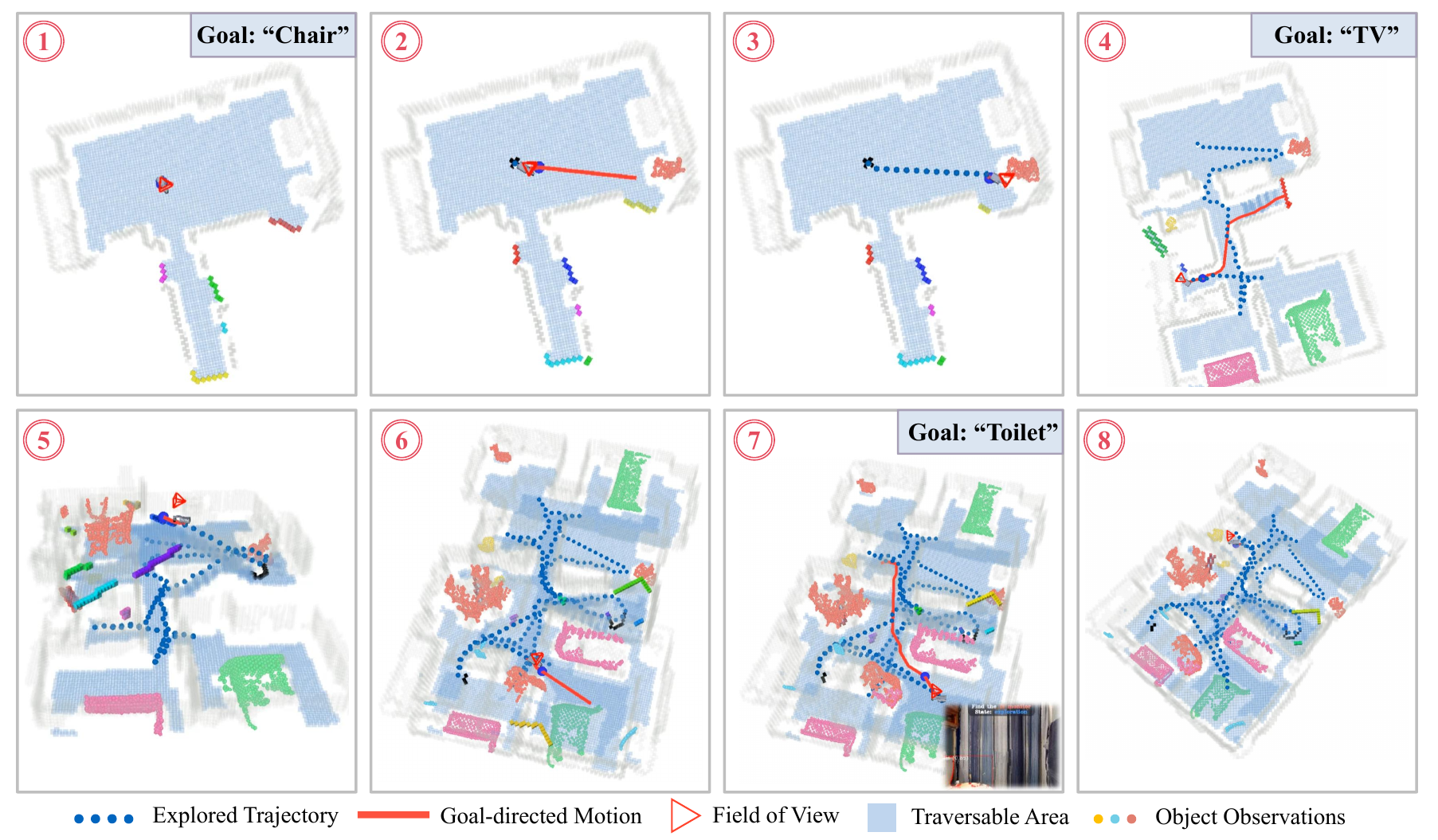}
    \caption{\textbf{Qualitative Cross-Floor-Required navigation on MFMON.}
    The episode contains three ordered goals, Chair $\rightarrow$ TV
    $\rightarrow$ Toilet, and requires one floor transition.}
    \label{fig:qualitative_cfr}
\end{figure}

\paragraph{Qualitative cross-floor analysis.}
Figure~\ref{fig:qualitative_cfr} visualizes a representative CFR episode with
three ordered goals, Chair $\rightarrow$ TV $\rightarrow$ Toilet. The agent
observes Toilet at step~75 before it becomes the active goal. Because the second
goal, TV, lies on another floor, the agent finishes same-floor search at
step~168, transitions to stair search, and reaches the upper floor at step~236,
where Toilet is observed again at step~335 while searching for TV. Once the goal
switches to Toilet, the consolidated experience is retrieved and grounded
through atom-ObjectMap binding for direct path planning, avoiding renewed
exploration.

\paragraph{Multimodal lifelong navigation.}
Table~\ref{tab:goat_main} follows the standard GOAT-Bench aggregation. On Val-Unseen, EvoNav achieves $30.77\%$ SR and $21.52\%$ SPL. Compared with TANGO, EvoNav has a 1.3-point lower SR but a 5.0-point higher SPL, and it outperforms the official Modular GOAT baseline by 4.3 SPL points. These results indicate that EvoNav more efficiently reuses maps and experience across subtasks, while leaving room for further improvement in multimodal goal completion.

\begin{table}[t]
\centering
\caption{Official GOAT-Bench Val-Unseen subtask-based, episode-macro results
(\%).}
\label{tab:goat_main}
\small
\begin{tabular}{lcc}
\toprule
Method & Subtask SR $\uparrow$ & Subtask SPL $\uparrow$ \\
\midrule
SenseAct-NN Monolithic & 12.3 & 6.8 \\
Modular CLIP on Wheels & 16.1 & 10.4 \\
Modular GOAT & 24.9 & 17.2 \\
SenseAct-NN Skill Chain & 29.5 & 11.3 \\
TANGO & \textbf{32.1} & 16.5 \\
\midrule
EvoNav (ours) & $30.8$ & $\textbf{21.5}$ \\
\bottomrule
\end{tabular}
\end{table}

\subsection{Ablation Studies}
\label{sec:ablations}

\begin{table}[t]
\centering
\caption{Component ablations on HM3D-MFMON (\%).
}
\label{tab:ablation_components}
\small
\begin{tabular}{lcccc}
\toprule
Configuration & SeqSR $\uparrow$ & SeqSPL $\uparrow$ & PR $\uparrow$ & PPL $\uparrow$ \\
\midrule
w/o Consolidation & 27.7 & 12.8 & 42.8 & 20.3 \\
w/o E2K evolution & 28.2 & 13.3 & 43.4 & 20.9 \\
w/o AGEP Prop. & 28.8 & 13.9 & 44.2 & 21.7 \\
w/o Feedback  & 29.4 & 14.6 & 45.3 & 22.8 \\
w/o Obj. binding & 27.0 & 11.9 & 41.7 & 19.2 \\
\midrule
EvoNav (full) & $\mathbf{30.3}$ & $\mathbf{15.5}$ & $\mathbf{48.3}$ & $\mathbf{24.2}$ \\
\bottomrule
\end{tabular}
\end{table}

As shown in Table~\ref{tab:ablation_components}, removing
atom-ObjectMap binding causes the largest degradation, suggesting that memory
retrieval alone is insufficient unless it is regrounded in current geometry.
Disabling online consolidation or E2K evolution reduces SeqSR by
approximately $2.6$ and $2.1$ points, respectively, supporting the complementary
roles of suppressing redundant observations and extracting reusable structural
knowledge. Graph propagation contributes indirect relational evidence, whereas
feedback calibration provides a smaller but consistent gain by suppressing
repeated false guidance.

\subsection{Memory Efficiency and Scaling}
\label{sec:memory_efficiency}

We define core persistent memory as the state used to store and retrieve experience across subtasks, excluding auxiliary geometric and planning maps. Table~\ref{tab:memory_efficiency}
summarizes core memory and access cost under a common per-episode convention.
3D-Mem and MemoryExplorer use snapshots as persistent units, LifelongCrossNav
uses feature-bearing voxels, and EvoNav stores atoms, experience links, and
evolved class knowledge. EvoNav retains only $16.16$ atoms on average,
corresponding to approximately $0.1$ MiB of core memory, with slow-update and query times of approximately $7$ ms and $2$ ms,
respectively. By comparison, the core memory of 3D-Mem,
MemoryExplorer, and LifelongCrossNav is approximately $10.2$, $87.4$, and at
least $156.2$ MiB per episode, respectively.

\begin{table}[t]
\centering
\caption{Core persistent-memory footprint and access cost per episode, measured on an NVIDIA A6000 GPU. }
\label{tab:memory_efficiency}
\small
\begin{tabular}{@{}lccc@{}}
\toprule
Method & Memory (MiB/ep.) & Update (s) & Query (s) \\
\midrule
3D-Mem
& 10.2 & 0.420 & 0.310 \\
MemoryExplorer
& 87.4 & 126.700 & 21.000 \\
LifelongCrossNav
& $156.2$ & 0.035 & 0.009 \\
\midrule
EvoNav (ours)
& \textbf{0.1} & \textbf{0.007} & \textbf{0.002} \\
\bottomrule
\end{tabular}
\end{table}

\subsection{Real-World Experiments}
\label{sec:real_world}

\begin{figure}[t]
    \centering
    \includegraphics[width=\linewidth]{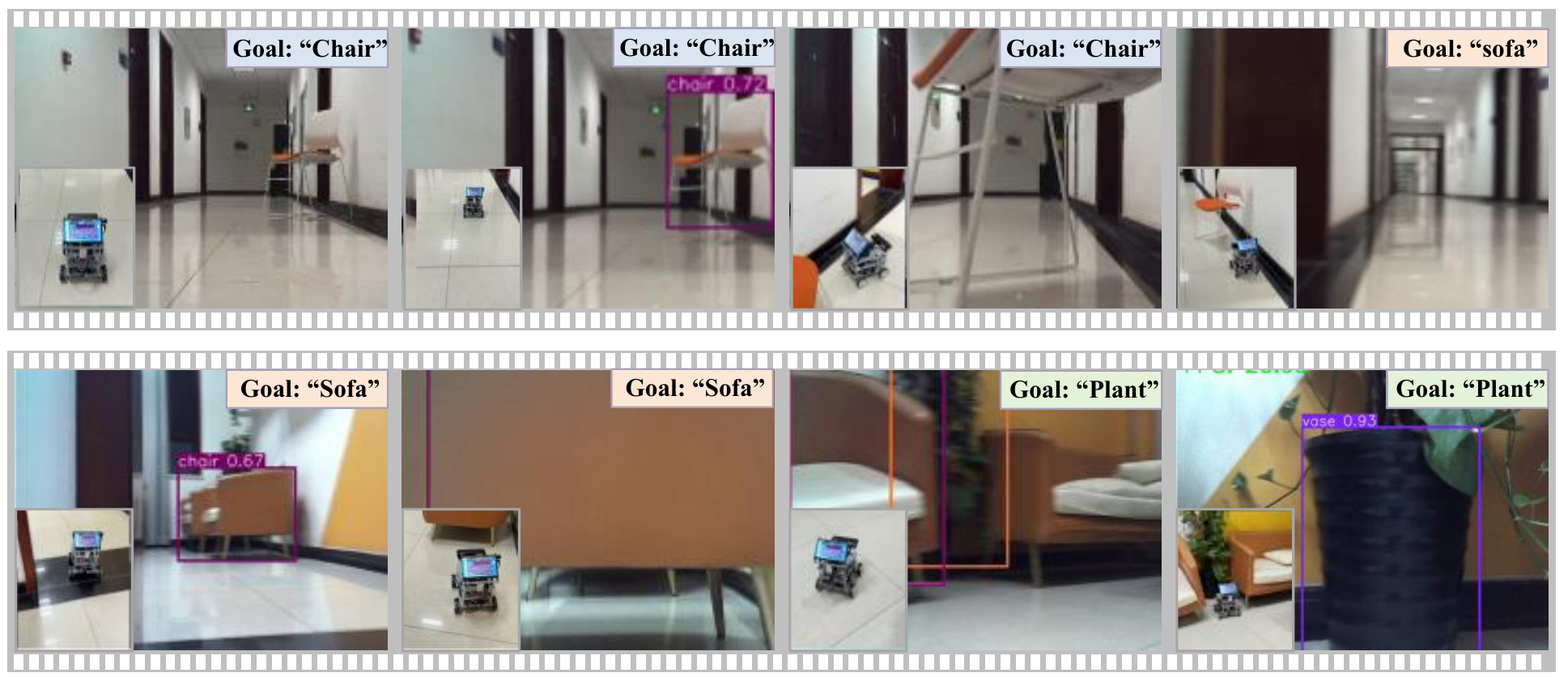}
    \caption{\textbf{Real-world navigation results.}
    Without additional tuning, EvoNav localizes and reaches target objects in a
    real indoor environment, demonstrating effective sim-to-real transfer.}
    \label{fig:real_world_navigation}
\end{figure}

We deploy EvoNav on a Wheeltec R550 mobile robot equipped with an Astra Pro Plus
RGB-D camera and an NVIDIA Jetson Orin NX Super 16GB module. The effective
sensing range is $0.6$-$5$ m. Open-vocabulary perception runs on a remote
workstation with an NVIDIA RTX A6000 GPU, while localization, mapping, and motion
control execute on board. The robot and workstation communicate over a local
network using ROS~2. On this platform, EvoNav successfully completes long-horizon navigation trials in real environments. Representative stages are
shown in Figure~\ref{fig:real_world_navigation}.

\section{Conclusion}

We presented Experience-Evolving Memory, a lightweight memory framework for lifelong embodied navigation. The method consolidates semantic observations, spatial relations, and behavioral transitions into sparse experience atoms, and progressively evolves repeated interactions into environment-specific structural knowledge. During navigation, learned experience provides goal-relevant search guidance, while the ObjectMap and traversability map ground memory in geometrically feasible actions. This design supports experience reuse across consecutive subtasks, regions, and floors without manually injected layout priors. Extensive experiments show that our method achieves strong path efficiency with a compact persistent-memory footprint, validating the effectiveness of experience evolution.
{
    \small
    \bibliographystyle{ieeenat_fullname}
    \bibliography{main}
}


\end{document}